\documentclass[runningheads]{llncs}

\usepackage{eccv}
\usepackage{eccvabbrv}
\usepackage{graphicx}
\usepackage{booktabs}
\usepackage{pifont}
\newcommand{\cmark}{\ding{51}}%
\newcommand{\xmark}{\ding{55}}%
\usepackage[accsupp]{axessibility}  

\usepackage{hyperref}
\usepackage{orcidlink}

\begin{document}

\title{ModuSeg: Decoupling Object Discovery and Semantic Retrieval for Training-Free Weakly Supervised Segmentation} 

\titlerunning{ModuSeg: Decoupling Object Discovery and Semantic Retrieval}
 
\author{Qingze He\inst{1}\orcidlink{0009-0005-0859-2770} \and
Fagui Liu\inst{1,2}\thanks{Corresponding authors.}\orcidlink{0000-0003-1135-4982} \and
Dengke Zhang\inst{1}\orcidlink{0009-0001-2941-0084} \and
Qingmao Wei\inst{1,2}\orcidlink{0000-0001-9982-3119} \and
Quan~Tang\inst{2}$^{\star}$\orcidlink{0000-0003-4011-6166}
}

\authorrunning{Q.~He et al.}

\institute{South China University of Technology, Guangzhou, China \and
Pengcheng Laboratory, Shenzhen, China \\
\email{202230430064@mail.scut.edu.cn, fgliu@scut.edu.cn, tangq@pcl.ac.cn}}

\maketitle

\begin{abstract}
  Weakly supervised semantic segmentation aims to achieve pixel-level predictions using image-level labels. Existing methods typically entangle semantic recognition and object localization, which often leads models to focus exclusively on sparse discriminative regions. Although foundation models show immense potential, many approaches still follow the tightly coupled optimization paradigm, struggling to effectively alleviate pseudo-label noise and often relying on time-consuming multi-stage retraining or unstable end-to-end joint optimization. To address the above challenges, we present ModuSeg, a training-free weakly supervised semantic segmentation framework centered on explicitly decoupling object discovery and semantic assignment. Specifically, we integrate a general mask proposer to extract geometric proposals with reliable boundaries, while leveraging semantic foundation models to construct an offline feature bank, transforming segmentation into a non-parametric feature retrieval process. Furthermore, we propose semantic boundary purification and soft-masked feature aggregation strategies to effectively mitigate boundary ambiguity and quantization errors, thereby extracting high-quality category prototypes. Extensive experiments demonstrate that the proposed decoupled architecture  better preserves fine boundaries without parameter fine-tuning and achieves highly competitive performance on standard benchmark datasets. Code is available at \url{https://github.com/Autumnair007/ModuSeg}.
\end{abstract}


\section{Introduction}
\label{sec:intro}

Weakly supervised semantic segmentation (WSSS) aims to achieve pixel-level predictions using image-level labels \cite{ahn2018learning,pinheiro2015image}, thereby substantially reducing dense annotation costs. Existing mainstream methods typically rely on class activation mapping \cite{zhou2016learning} to provide initial localization cues. However, these methods inherently couple semantic recognition and object localization within a single classification objective. Consequently, as classification networks naturally tend to focus on the most discriminative local regions \cite{chen2023extracting,wu2024dino,zhang2025exploring}, the generated activation maps often cover only sparse object fragments. To rectify these incomplete initial regions, traditional paradigms typically rely on time-consuming multi-stage network retraining \cite{ru2022learning} or struggle with the instability of complex end-to-end joint training. This coupled design increases training complexity and restricts the geometric precision of the final segmentation masks.

The recent advancement of visual foundation models \cite{radford2021learning} provides a new opportunity to address the above dilemma. Class-agnostic segmentation models have demonstrated exceptional boundary perception capabilities, while vision-language and self-supervised models have learned highly expressive dense semantic descriptions from massive data \cite{zhou2022extract}. Despite their potential, many existing weakly supervised segmentation approaches still follow the conventional path of coupled optimization, attempting to integrate foundation model priors through network fine-tuning or end-to-end distillation \cite{lin2023clip,yang2025exploring,bi2026ssr}. Such approaches often struggle to circumvent the interference of inherent noise in weakly supervised pseudo-labels, underutilizing the true capabilities of foundation models.

\begin{figure}[!t]
  \centering
  \includegraphics[trim=20mm 70mm 10mm 50mm, clip, width=\linewidth]{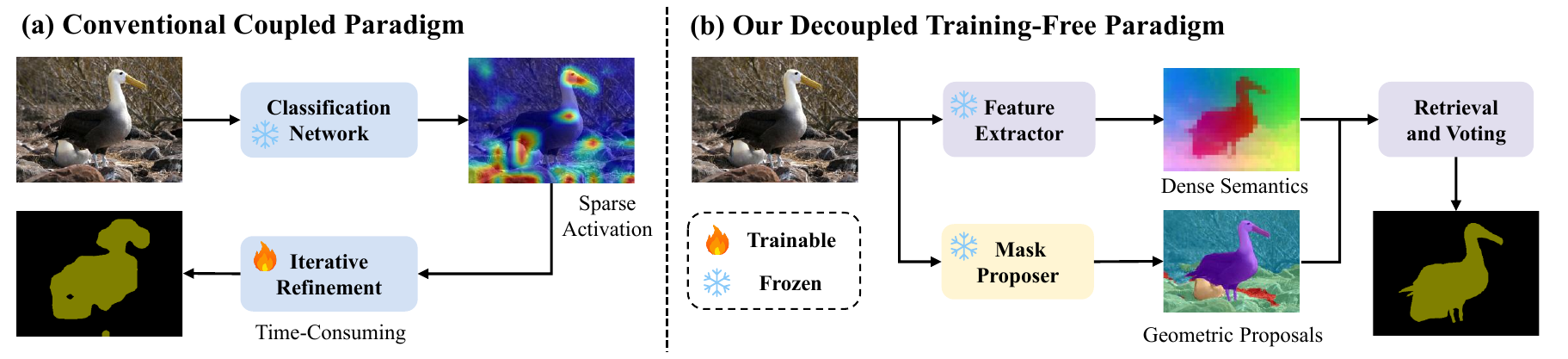}
  \caption{\textbf{Comparison of paradigms.} Unlike conventional coupled methods that rely on iterative refinement, our decoupled approach directly assigns semantics to geometric proposals via a training-free retrieval mechanism.}
  \label{fig:motivation}
\end{figure}

To address the limitations of conventional coupled paradigms illustrated in \cref{fig:motivation} (a), we propose ModuSeg, a novel training-free weakly supervised semantic segmentation framework shown in \cref{fig:motivation} (b). The core insight lies in the explicit decoupling of object discovery and semantic assignment. We delegate the localization task to an off-the-shelf, high-performance general mask proposer, which provides geometric proposals with reliable boundaries. Simultaneously, we utilize semantic foundation models to construct an offline feature bank. By transforming the segmentation task into a non-parametric feature retrieval process tailored for geometric proposals, our method better preserves the fine physical boundaries of objects while assigning semantics. This decoupling strategy facilitates a highly modular paradigm, enabling our approach to bypass time-consuming retraining and seamlessly integrate continuously evolving feature extractors.

To mitigate the structural alignment errors and boundary noise inherent in weakly supervised pseudo-labels, we further design two targeted modules to enhance the feature bank construction. We propose a semantic boundary purification strategy to alleviate feature mixing by stripping away ambiguous edge regions, thereby improving the semantic purity of the extracted features. Furthermore, we introduce a soft-masked feature aggregation mechanism to pool visual embeddings via soft weights, which alleviates the noise caused by spatial hard quantization. These modules operate in conjunction to extract discriminative semantic representations for the subsequent retrieval process.

The main contributions of this work are summarized as follows:
\begin{itemize}
\item We propose the explicit decoupling of object discovery and semantic retrieval, diverging from traditional approaches constrained by coupled optimization (whether multi-stage or end-to-end), and constructing an efficient, training-free modular segmentation framework.
\item We introduce the semantic boundary purification and soft-masked feature aggregation modules during the feature bank construction stage. These modules mitigate boundary ambiguity and quantization errors, improving the quality of feature prototypes.
\item ModuSeg achieves highly competitive performance on the primary PASCAL VOC and MS COCO benchmarks, while auxiliary evaluations on ADE20K and Cityscapes further support generalizability across diverse datasets.
\end{itemize}


\section{Related Work}

\subsection{Weakly Supervised Semantic Segmentation}
The fundamental challenge in WSSS lies in the inherent coupling between semantic classification and object localization. Traditional approaches rely on Class Activation Maps (CAMs) \cite{zhou2016learning}, which inevitably focus on sparse discriminative object parts rather than the complete object extent. To rectify these incomplete parts, earlier methods such as AffinityNet \cite{ahn2018learning} require complex iterative refinement and computationally expensive retraining of a final segmentation network.

Recent advancements leverage stronger priors from Vision Transformers (ViT), Vision-Language Models (VLMs) and class-agnostic segmentation models. Methods such as ToCo \cite{ru2023token} and MCTformer+ \cite{xu2024mctformer+} utilize self-attention mechanisms to encourage integral object activation. Meanwhile, CLIP-based approaches have emerged as a dominant paradigm. WeCLIP \cite{zhang2024frozen} and WeakCLIP \cite{zhu2025weakclip} freeze the CLIP backbone to extract dense knowledge. SAM-based methods such as SEPL \cite{chen2023segment} and S2C \cite{kweon2024sam} exploit class-agnostic mask priors to improve CAM or pseudo-label quality. However, these methods rely on a coupled optimization paradigm, forcing models to learn dense predictions from inherently noisy pseudo-supervision. In contrast, our work decouples object discovery and semantic retrieval by formulating segmentation as a training-free proposal-retrieval process.

\subsection{Vision Foundation Models}
The evolution of foundation models suggests a new philosophy in which specialized models achieve superior performance in their respective decoupled domains. On the semantic front, dense vision foundation models such as DINOv2/v3 \cite{oquab2023dinov2,simeoni2025dinov3} and C-RADIOv4 \cite{ranzinger2026c} provide robust pixel-wise semantic descriptors that surpass supervised baselines. On the geometric front, class-agnostic models such as EntitySeg \cite{Qi_2023_ICCV} and SAM 2 \cite{ravi2024sam} have effectively addressed the localization problem by providing high-fidelity boundaries independent of semantic categories. 

Open-Vocabulary Semantic Segmentation (OVSS) assigns pixel labels using textual category descriptions, including categories unseen during training, covering training-free \cite{shi2025harnessing,xuan2025reme,zhang2025corrclip} and supervised \cite{cho2024cat,li2025mask} methods. In contrast, standard WSSS learns from image-level labels for fixed category sets. ModuSeg follows this WSSS setting: image-level labels constrain feature-bank construction rather than define open-vocabulary queries. While OVSS models provide strong semantic priors, directly using their predictions for WSSS risks hallucinating absent categories (\ie, false positives). Therefore, ModuSeg does not pursue open-vocabulary generalization; it integrates pre-trained foundation models into an image-level constrained WSSS framework, where proposal-level retrieval assigns semantics from the feature bank.

\subsection{Prototype Learning}
Prototype learning reformulates segmentation by representing categories as feature centroids rather than parametric weights, employing nearest-neighbor retrieval or feature reconstruction for dense prediction \cite{zhou2022rethinking,tang2025rethinking}. In WSSS, this paradigm is adapted to bridge the modality gap. For instance, SSR \cite{bi2026ssr} aligns visual features with learnable text prototypes to reduce class confusion, while ExCEL \cite{yang2025exploring} enriches text prototypes via LLMs for fine-grained patch alignment. However, these approaches typically rely on intricate optimization processes for mask refinement. In contrast, we formulate segmentation as a retrieval task. By leveraging a purified feature bank derived from frozen foundation models, our method ensures robust semantic assignment without parameter tuning.



\section{Methodology}
\label{sec:methodology}

\subsection{Method Overview}

\begin{figure}[!t]
  \centering
  \includegraphics[trim=5mm 10mm 5mm 10mm, clip, width=\textwidth]{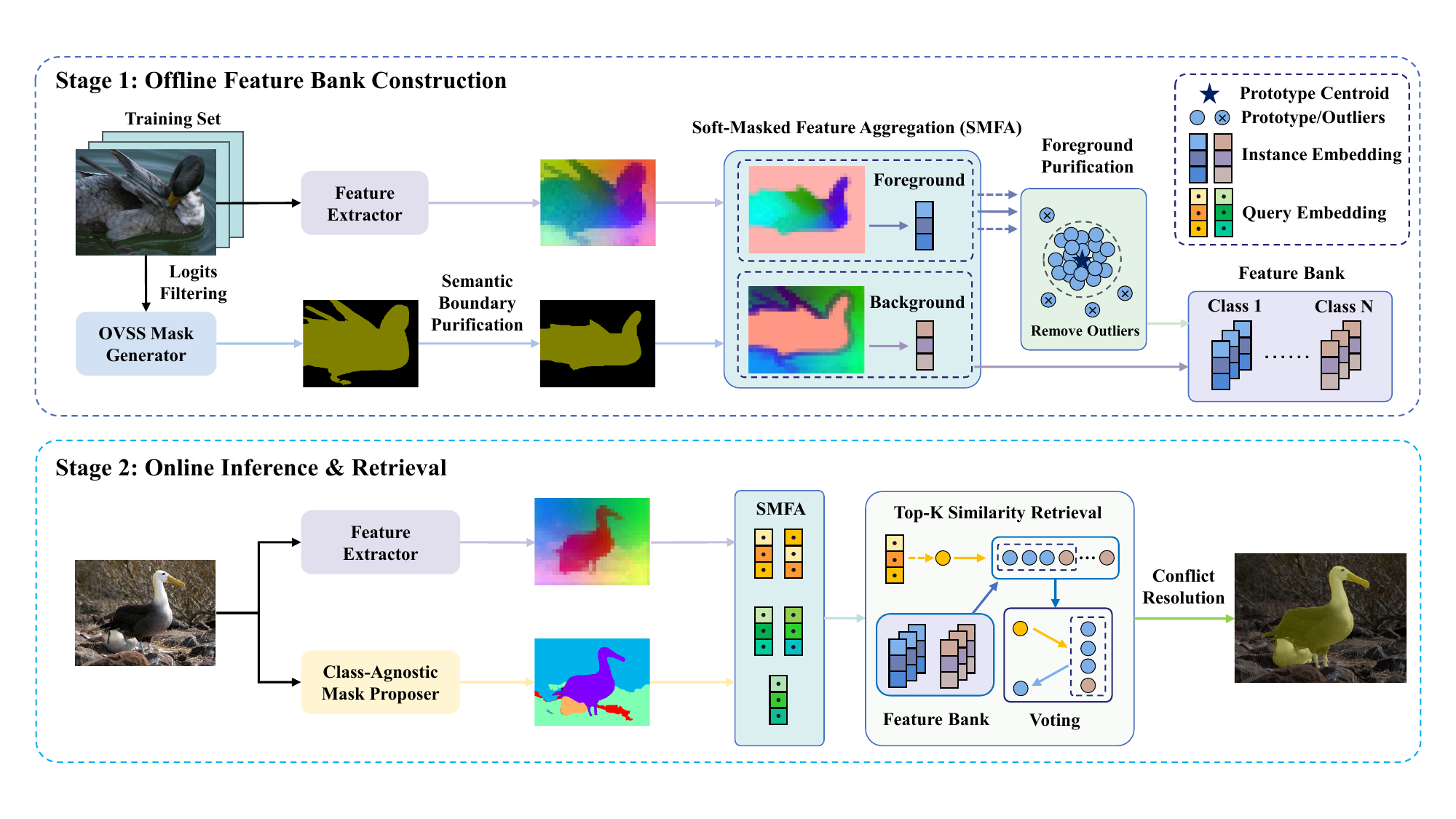}
  \caption{\textbf{Overview of our proposed ModuSeg.} The framework operates in two stages: (1) Offline Feature Bank Construction, where we refine semantic boundaries to extract representative prototypes via soft-masked aggregation; and (2) Online Inference \& Retrieval, which assigns semantics to class-agnostic proposals through a retrieval-based voting mechanism, effectively decoupling localization from classification.}
  \label{fig:model_overview}
\end{figure}

As illustrated in \cref{fig:model_overview}, ModuSeg decouples object discovery from classification to achieve training-free semantic segmentation. The pipeline consists of: (1) Robust feature bank construction. We utilize image-level supervision to generate initial masks, which are processed via semantic boundary purification to mitigate feature mixing at object edges. These refined regions are converted into prototype vectors using soft-masked feature aggregation and filtered to enhance intra-class compactness. (2) Class-agnostic object discovery. During inference, we utilize a frozen, class-agnostic mask proposer to extract geometric proposals across diverse object categories. (3) Retrieval-augmented semantic assignment. We extract query embeddings for each proposal via soft masking and perform Top-K similarity retrieval against the offline bank. A majority voting strategy determines the class label, followed by confidence-priority rasterization to resolve spatial conflicts and produce the final segmentation.

\subsection{Stage 1: Constructing a Robust Feature Bank}
\label{sec:stage1}

The efficacy of a retrieval-based segmentation framework hinges on the representational quality of its feature prototypes. In this stage, we focus on extracting semantically pure and discriminative representations from the training images to construct a high-fidelity feature bank.


\subsubsection{Mask Generation with Semantic Boundary Purification}
\label{sec:sbp}

To initialize the feature bank, we utilize image-level supervision to generate candidate regions. Let $\mathcal{D} = \{(I_i, \mathcal{Y}_i)\}_{i=1}^N$ denote the training dataset, where $I_i$ denotes the $i$-th image, $\mathcal{Y}_i$ is the set of ground-truth class labels and $N$ denotes the total number of images. We employ CorrCLIP \cite{zhang2025corrclip} as our base mask generator $\mathcal{G}_{mask}$.

\textbf{Image-Level Logits Filtering.} While CorrCLIP provides a strong baseline, applying it directly can introduce false positives (\ie, hallucinated classes). To ensure semantic fidelity, we modify the generation process by incorporating an image-level logits filtering mechanism. Specifically, before the final prediction, we mask the class logits corresponding to categories not present in $\mathcal{Y}_i$ by setting their values to $-\infty$. This constraint forces the model to assign pixels only to the classes actually present in the image or to the background. Formally, we generate the initial map $S_i$ as follows:
\begin{equation}
    S_i = \mathcal{G}_{mask}(I_i, \mathcal{Y}_i).
\end{equation}
Here, $\mathcal{G}_{mask}$ denotes the mask generator conditioned on the image-level labels $\mathcal{Y}_i$, ensuring that $S_i$ contains only valid categories.

\textbf{Semantic Boundary Purification (SBP).} Even with semantically correct labels, extracting features using ViT encounters two challenges regarding boundary fidelity: (1) \textbf{Structural Patch-Pixel Misalignment}, where the fixed-grid processing of ViT (\eg, $16 \times 16$ patches) inevitably leads to patches at boundaries encoding mixed foreground-background representations; and (2) \textbf{Inherent Mask Noise}, as pseudo-masks generated from weak supervision often exhibit uncertainty and irregularity along object edges. 

To address these challenges simultaneously, we propose the Semantic Boundary Purification (SBP) strategy. We first derive the raw binary mask for class $c$ from the initial map $S_i$ as $m_c = \mathbb{I}(S_i = c)$, where $\mathbb{I}(\cdot)$ denotes the indicator function. Instead of utilizing this raw mask directly, we apply $t$ iterations of strict morphological erosion to explicitly discard ambiguous boundary regions:
\begin{equation}
    m_c^{pure} = \text{Erode}^{(t)}(m_c, B_k).
\end{equation}
In this equation, $m_c^{pure}$ denotes the resulting purified mask, $B_k$ represents a structuring element with a kernel size of $k$, and $t$ denotes the number of iterations. This operation serves a twofold purpose: it eliminates feature mixing by ensuring that the extracted patches fall strictly within the object interior, and it filters mask noise by pruning the unreliable edge pixels. By sacrificing geometric completeness at the boundaries, SBP ensures that the subsequent feature extraction focuses solely on the semantic core of the object, thereby maximizing the purity of the feature bank.


\subsubsection{Modular Feature Extraction}
\label{sec:smfa}

Given the purified masks $m_c^{pure}$, we extract their corresponding visual embeddings. A naive approach is to downsample the mask to the feature grid size using nearest-neighbor interpolation. However, this hard quantization exacerbates spatial quantization noise, assigning binary labels to patches that may only be partially covered by the mask.

To achieve a more precise representation, we employ Soft-Masked Feature Aggregation (SMFA). Let $F \in \mathbb{R}^{h \times w \times D}$ be the feature map extracted from the frozen backbone $\Phi$, where $h$, $w$, and $D$ denote the height, width, and channel dimension of the feature grid, respectively. We project the high-resolution purified mask $m_c^{pure}$ onto the low-resolution feature grid via area-based interpolation, resulting in a soft assignment map $W \in [0, 1]^{h \times w}$. The value $W_{x,y}$ quantifies the semantic contribution of patch $(x,y)$ to the object class $c$.

The instance-specific prototype vector $v_c \in \mathbb{R}^D$ is computed as a weighted aggregate:
\begin{equation}
    v_c = \text{Normalize}\left( \frac{\sum_{x,y} W_{x,y} \cdot F_{x,y}}{\sum_{x,y} W_{x,y} + \epsilon} \right).
\end{equation}
Here, $\epsilon$ is a small constant to prevent division by zero. This soft-weighting mechanism ensures that patches with higher mask coverage contribute more to the final prototype, while marginal patches are naturally suppressed. This design effectively decouples mask generation from feature extraction, rendering our framework model-agnostic and compatible with various ViT architectures.


\subsubsection{Prototype-based Feature Purification}

Since pseudo-masks are inherently noisy, the initial feature bank inevitably contains outliers—features corresponding to misaligned regions or ambiguous visual patterns. To enhance intra-class compactness, we introduce a prototype-based purification mechanism.

\textbf{The Cluster Assumption.} Our strategy relies on the Cluster Assumption: in the embedding space, reliable features of the same semantic category tend to form a dense, unimodal cluster, whereas noisy samples are sparsely distributed in the periphery.

\textbf{Selective Outlier Rejection.} For each semantic class $c$, we aggregate all extracted feature vectors $\mathcal{V}_c$ to compute a global class prototype $\mu_c$, which serves as the robust centroid of the distribution: $\mu_c = \frac{1}{|\mathcal{V}_c|} \sum_{v \in \mathcal{V}_c} v$.

We then measure the semantic deviation of each instance $v$ via its Euclidean distance from the prototype $d_v = \| v - \mu_c \|_2$. To refine the feature bank, we discard the top $\alpha\%$ of samples with the largest deviations:
\begin{equation}
    \mathcal{V}_c^{clean} = \left\{ v \in \mathcal{V}_c \;\middle|\; d_v \le \text{Quantile}\left(1-\frac{\alpha}{100}, \{d_{v'} \mid v' \in \mathcal{V}_c\}\right) \right\}.
\end{equation}

Notably, this purification is applied exclusively to foreground classes. While foreground objects generally exhibit high intra-class similarity (\eg, ``cars'' share common visual traits), the ``background'' class is inherently multimodal, encompassing diverse entities such as sky, roads, and vegetation. Filtering background features based on a single centroid would reduce the diversity of the negative samples, impairing the model's ability to distinguish objects from complex environments. Therefore, we retain all background embeddings to maintain a comprehensive representation of the visual context.


\subsection{Stage 2: Inference via Semantic Retrieval}
\label{sec:stage2}

Unlike traditional WSSS paradigms that require a resource-intensive final segmentation network retraining on pseudo-labels, our approach explicitly decouples the task into two independent, training-free components: Class-Agnostic Localization and Retrieval-based Classification. This non-parametric design ensures high-quality boundary delineation using off-the-shelf geometric priors.


\subsubsection{Localization via Class-Agnostic Object Discovery}
\label{sec:localization}

Traditional WSSS methods often struggle with ``boundary ambiguity'' or ``over-activation'' because classification networks lack explicit geometric constraints during the localization phase. To address these geometric deficiencies, we utilize specialized object discovery models as a Class-Agnostic Mask Proposer.

Formally, given a test image $I_{\text{test}}$, we employ a generalized segmentation model $\Psi$ to extract a set of potential object proposals $\mathcal{P}$:
\begin{equation}
    \mathcal{P} = \Psi(I_{\text{test}}) = \{ (m_p, s_p^{obj}) \}_{p=1}^M.
\end{equation}
Here, $M$ is the total number of proposals, and $m_p \in \{0, 1\}^{H \times W}$ denotes the binary mask of the $p$-th proposal, where $H$ and $W$ are scalars representing the height and width of the input image, respectively. $s_p^{obj} \in [0, 1]$ represents its objectness score, indicating the confidence of being a valid object.

\textbf{Training-Inference Asymmetry.} A critical distinction in our framework lies in handling mask boundaries across stages. While we apply SBP during the feature bank construction (\cref{sec:sbp}) to ensure feature purity, we utilize raw, un-eroded proposals during inference. This asymmetry is intentional: training prioritizes precision, whereas inference prioritizes completeness by using raw proposals to preserve fine-grained geometric details that SBP otherwise discards. Furthermore, this design accommodates the inherent limitations of class-agnostic proposers. Unlike pristine ground-truth annotations, they typically over-partition scenes into fragmented entities based on low-level cues (as diagnosed in \cref{sec:deep_analysis}). Consequently, eroding these already fragmented inference proposals severely diminishes their representational capacity and degrades matching accuracy, even when they are strictly used for feature retrieval. Thus, relying on raw proposals during inference is essential to prevent exacerbating the semantic gap.

To ensure computational efficiency, we filter out low-quality proposals. We discard candidates with an objectness score below a threshold $\tau_{obj}$:
\begin{equation}
    \mathcal{P}_{valid} = \{ m_p \mid (m_p, s_p^{obj}) \in \mathcal{P}, s_p^{obj} \ge \tau_{obj} \}.
\end{equation}


\subsubsection{Retrieval-Augmented Semantic Assignment}
\label{sec:retrieval}

Once a valid geometric proposal $m_p \in \mathcal{P}_{valid}$ is obtained, the subsequent task is to assign a semantic label to it. We address this semantic assignment via a non-parametric retrieval process from our offline feature bank $\mathcal{B} = \bigcup_c \mathcal{V}_c^{clean}$.

\textbf{Query Extraction.} For each proposal $m_p$, we extract its visual embedding $q_p$ using the same frozen backbone $\Phi$ used in Stage 1. To ensure consistency between training and inference representations, we apply the same SMFA strategy (\cref{sec:smfa}) to pool the feature maps within the region defined by $m_p$.

\textbf{Top-K Similarity Retrieval.} Rather than relying on a single nearest neighbor, we identify the $K$ most similar visual embeddings within the feature bank $\mathcal{B}$ to provide a robust semantic context for $q_p$. Let $\mathcal{N}_K(q_p) = \{(v_j, y_j)\}_{j=1}^K$ denote the set of retrieved Top-K candidates, where $v_j$ is the feature vector and $y_j$ is its corresponding class label. The similarity between the query $q_p$ and each retrieved feature $v_j$ is quantified using cosine similarity: 
\begin{equation}
\text{sim}(q_p, v_j) = \frac{q_p \cdot v_j}{\|q_p\| \|v_j\|}.
\end{equation}

\textbf{Hierarchical Semantic Voting.} To determine the class label $\hat{c}_p$, we propose a hierarchical voting strategy that prioritizes neighborhood consensus over raw similarity magnitude. This strategy mitigates the influence of outliers that may possess high similarity scores despite being semantically irrelevant.

First, we calculate the vote count for each class $c$ present in the neighborhood:
\begin{equation}
    \text{Vote}(c) = \sum_{(v_j, y_j) \in \mathcal{N}_K(q_p)} \mathbb{I}(y_j = c).
\end{equation}

We assign class labels by majority vote $\hat{c}_p = \arg\max_{c} \text{Vote}(c)$. In cases where multiple classes receive the same number of votes, we break ties by selecting the class with the highest cumulative similarity among the supporting samples. This strategy effectively treats classification as a lexicographical optimization problem: first prioritizing consensus count, and then similarity magnitude.

\textbf{Confidence Estimation.} To facilitate the subsequent conflict resolution step (\cref{sec:conflict}), we compute a semantic confidence score $s_p^{sem}$ for the proposal. This score is defined as the average cosine similarity of the neighbors belonging to the winning class $\hat{c}_p$:
\begin{equation}
    s_p^{sem} = \frac{1}{\text{Vote}(\hat{c}_p)} \sum_{(v_j, y_j) \in \mathcal{N}_K(q_p), y_j = \hat{c}_p} \text{sim}(q_p, v_j).
\end{equation}
This metric effectively quantifies the reliability of the prediction by measuring the density of the feature manifold around the query.


\subsubsection{Conflict Resolution and Rasterization}
\label{sec:conflict}

Given the dense generation of object proposals, the candidate set inherently contains overlapping proposals. We employ a two-step strategy to consolidate these candidates into a non-overlapping semantic segmentation map.

First, we apply Class-Specific Non-Maximum Suppression (NMS) to eliminate redundant detections. Specifically, we perform NMS independently for each predicted class $\hat{c}_p$ and within each group using an Intersection-over-Union (IoU) threshold $\tau_{nms}$. This process removes duplicate predictions while preserving valid spatial overlaps between distinct categories (\eg, a person riding a bicycle).

Second, to resolve pixel-level conflicts between categories, we employ confidence priority rasterization. We rank the remaining candidates in descending order based on their semantic confidence scores $s_p^{sem}$ (derived in \cref{sec:retrieval}). The final semantic map $S_{final}$ is constructed by sequentially assigning pixels to the highest-ranking candidate:
\begin{equation}
    S_{final}(u, v) = \hat{c}_p \quad \text{if } (u, v) \in m_p \text{ and } S_{final}(u, v) = \emptyset.
\end{equation}
Here, $S_{final}(u, v) = \emptyset$ denotes that the pixel at $(u, v)$ is currently unassigned. This ``first-come-first-served'' mechanism ensures that high-confidence predictions take precedence, handling occlusions and resolving boundary ambiguities.


\section{Experiments}
\label{sec:experiments}


\subsection{Experimental Settings}
\label{sec:exp_settings}

\textbf{Datasets and Metrics.} We primarily evaluate ModuSeg on two standard WSSS benchmarks: PASCAL VOC 2012 \cite{everingham2015pascal} (10,582 train, 1,449 val, 1,456 test images) and MS COCO 2014 \cite{lin2014microsoft} (82,081 train, 40,137 val images). Accordingly, our main experimental results are reported on VOC and COCO. ADE20K \cite{zhou2017scene} and Cityscapes \cite{cordts2016cityscapes} are used for generalization analysis. We adopt mean Intersection-over-Union (mIoU) as the primary metric.

\textbf{Implementation Details.} Our framework is implemented in PyTorch \cite{paszke2019pytorch} using the pre-trained C-RADIOv4-SO400M \cite{ranzinger2026c} backbone and CorrCLIP \cite{zhang2025corrclip} for pseudo-mask generation. To purify the feature bank, we perform morphological erosion using a $3\times3$ kernel for 20 iterations and discard the 25\% of features with the largest prototype distances. During inference, we utilize EntitySeg \cite{Qi_2023_ICCV} for mask proposals and set the retrieval neighbor count to $K=25$. For the retrieval backend, we employ the FAISS library \cite{douze2025faiss}, adopting a deterministic flat index for PASCAL VOC to ensure reproducibility, while utilizing a GPU-accelerated inverted file index (IVF) for MS COCO to accelerate inference on the large-scale dataset. More details are provided in the supplementary material.

\subsection{Comparisons with State-of-the-Art Methods}

\begin{table}[!t]
    \centering
    \caption{\textbf{Segmentation comparison with state-of-the-art methods on VOC and COCO.} Train: \cmark=Required, \xmark=Training-Free.}
    \label{tab:sota}
    \small 
    \setlength{\tabcolsep}{5pt}
    \begin{tabular}{l|c|cc|c} 
    \toprule
    Method & Train & \multicolumn{2}{c|}{VOC (mIoU)} & COCO (mIoU) \\ 
    & & Val & Test & Val \\
    \midrule
    \multicolumn{5}{l}{\textit{\textbf{Single-stage methods}}} \\
    ToCo {\scriptsize (CVPR'23)} \cite{ru2023token}           & \cmark & 71.1 & 72.2 & 42.3 \\
    DuPL {\scriptsize (CVPR'24)} \cite{wu2024dupl}            & \cmark & 73.3 & 72.8 & 44.6 \\
    WeCLIP {\scriptsize (CVPR'24)} \cite{zhang2024frozen}     & \cmark & 76.4 & 77.2 & 47.1 \\
    PCRE {\scriptsize (CVPR'25)} \cite{xu2025weakly}          & \cmark & 75.5 & 75.9 & 47.2 \\
    MoRe {\scriptsize (AAAI'25)} \cite{yang2025more}          & \cmark & 76.4 & 75.0 & 47.4 \\
    ExCEL {\scriptsize (CVPR'25)} \cite{yang2025exploring}    & \cmark & 78.4 & 78.5 & 50.3 \\
    TokenMasking {\scriptsize (ICCV'25)} \cite{hanna2025know} & \cmark & 72.7 & 73.5 & 43.2 \\
    SSR {\scriptsize (AAAI'26)} \cite{bi2026ssr}              & \cmark & 79.5 & 79.6 & 50.6 \\
    \midrule
    \multicolumn{5}{l}{\textit{\textbf{Multi-stage methods}}} \\
    CLIP-ES {\scriptsize (CVPR'23)} \cite{lin2023clip}            & \cmark & 73.8 & 73.9 & 45.4 \\
    MCTformer+ {\scriptsize (TPAMI'24)} \cite{xu2024mctformer+}   & \cmark & 74.0 & 73.6 & 45.2 \\
    CPAL {\scriptsize (CVPR'24)} \cite{tang2024hunting}           & \cmark & 74.5 & 74.7 & 46.8 \\
    S2C {\scriptsize (CVPR'24)} \cite{kweon2024sam}               & \cmark & 78.2 & 77.5 & 49.8 \\
    WeakCLIP {\scriptsize (IJCV'25)}  \cite{zhu2025weakclip}      & \cmark & 74.0 & 73.8 & 47.4 \\
    POT {\scriptsize (CVPR'25)}  \cite{wang2025pot}               & \cmark & 76.1 & 76.7 & 47.9 \\ 
    OTPL {\scriptsize (ICCV'25)} \cite{wang2025class}             & \cmark & 76.4 & 76.8 & 47.6 \\ 
    VPL {\scriptsize (AAAI'25)}  \cite{xu2025toward}              & \cmark & 79.3 & 79.0 & 49.8 \\
    \textbf{ModuSeg (Ours)}  & \textbf{\xmark} & \textbf{86.3} & \textbf{86.6} & \textbf{56.7} \\ 
    \bottomrule
    \end{tabular}
\end{table}

\begin{figure}[!t]
  \centering
  \includegraphics[width=\linewidth]{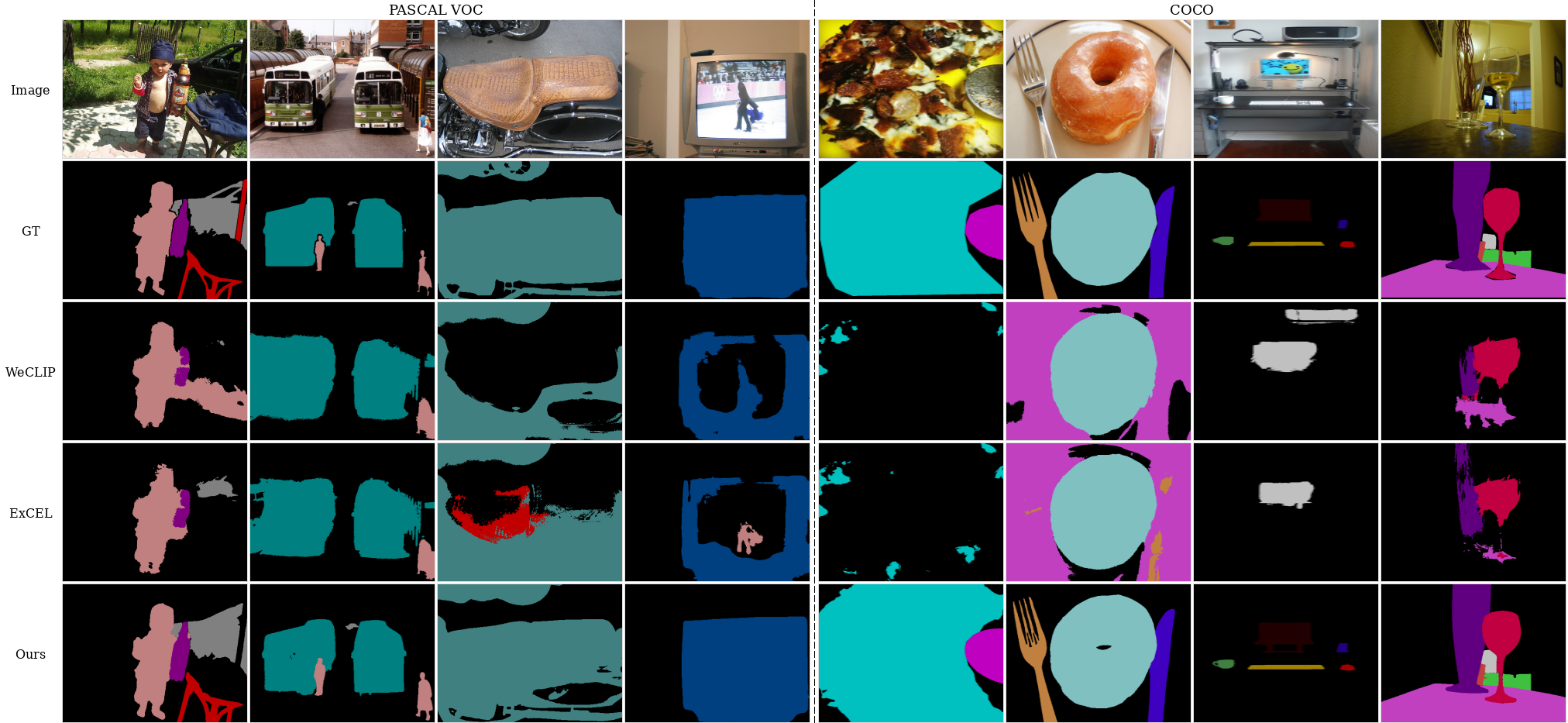}
  \caption{\textbf{Qualitative comparison on VOC and COCO.} ModuSeg demonstrates superior object completeness and boundary adherence compared to WeCLIP and ExCEL. Our method successfully delineates fine-grained details while eliminating the semantic inconsistencies and noise observed in competing approaches.}
  \label{fig:vis_comp}
\end{figure}

\textbf{Performance of Semantic Segmentation.} \cref{tab:sota} demonstrates the superior performance of ModuSeg. Recent state-of-the-art methods leverage foundation models for dense prediction, including VLM-based approaches such as WeCLIP and ExCEL, and SAM-based methods like S2C that employ SAM to refine CAMs during training. However, these coupled optimization strategies often suffer from noisy activation seeds or the coarse spatial resolution inherent in CLIP features. 
By decoupling object discovery from semantic retrieval, ModuSeg achieves \textbf{86.3\%} and \textbf{86.6\%} mIoU on VOC val and test sets, and \textbf{56.7\%} on COCO, surpassing the previous best-performing methods by 6.8\%, 7.0\% and 6.1\%, respectively. These results indicate that combining pretrained geometric and semantic priors through a decoupled proposal-retrieval pipeline is highly effective for WSSS. Qualitative comparisons in \cref{fig:vis_comp} further substantiate that our paradigm produces integral masks with precise boundaries, effectively suppressing the fragmentation and noisy artifacts inherent to baseline methods. Additional qualitative visualizations and interpretability analyses are provided in the supplementary material.

\begin{table}[!t]
    \centering
    \caption{\textbf{Comparison of initial seed quality (CAMs or pseudo-masks) on VOC Train set.} Since ModuSeg generates masks directly, we evaluate our generated pseudo-masks against the CAM seeds of state-of-the-art methods. \textbf{Sup.} denotes supervision: Image-level ($\mathcal{I}$), CLIP ($\mathcal{L}$), DINO ($\mathcal{D}$). \textbf{Arch.}: ViT (V), ResNet (R), DeiT (D).}
    \label{tab:mask_quality}
    \small
    \setlength{\tabcolsep}{6pt}
    \begin{tabular}{l|c|c|c}
    \toprule
    Method & Sup. & Arch. & VOC Train (mIoU) \\
    \midrule
    CLIP-ES {\scriptsize (CVPR'23)} \cite{lin2023clip}     & $\mathcal{I} + \mathcal{L}$ & V & 70.8 \\
    ToCo {\scriptsize (CVPR'23)} \cite{ru2023token}        & $\mathcal{I}$ & V & 73.6 \\
    CPAL {\scriptsize (CVPR'24)} \cite{tang2024hunting}    & $\mathcal{I} + \mathcal{L}$ & R & 71.9 \\
    CTI  {\scriptsize (CVPR'24)} \cite{yoon2024class}      & $\mathcal{I}$ & D & 69.5 \\
    SeCo {\scriptsize (CVPR'24)} \cite{yang2024separate}   & $\mathcal{I}$ & V & 74.8 \\
    DuPL {\scriptsize (CVPR'24)} \cite{wu2024dupl}         & $\mathcal{I}$ & V & 76.0 \\
    WeCLIP {\scriptsize (CVPR'24)} \cite{zhang2024frozen}  & $\mathcal{I} + \mathcal{L}$ & V & 75.4 \\
    WeakCLIP {\scriptsize (IJCV'25)} \cite{zhu2025weakclip}& $\mathcal{I} + \mathcal{L}$ & V & 61.7 \\
    VPL {\scriptsize (AAAI'25)}  \cite{xu2025toward}       & $\mathcal{I} + \mathcal{L}$ & V & 77.8 \\
    ExCEL {\scriptsize (CVPR'25)} \cite{yang2025exploring} & $\mathcal{I} + \mathcal{L}$ & V & 78.0 \\
    SSR {\scriptsize (AAAI'26)} \cite{bi2026ssr}           & $\mathcal{I} + \mathcal{L} + \mathcal{D}$ & V & 78.7 \\
    \midrule
    \textbf{ModuSeg (Ours)} & $\mathcal{I}+\mathcal{L}+\mathcal{D}$ & V & \textbf{78.8} \\
    \bottomrule
    \end{tabular}
\end{table}

\textbf{Evaluation of Initial Seed Quality.} \cref{tab:mask_quality} assesses the quality of initial seed on the VOC training set. By integrating image-level label filtering into the CorrCLIP backbone, our approach achieves \textbf{78.8\%} mIoU. This result establishes a new benchmark for seed generation, outperforming ExCEL by 0.8\% and effectively suppressing hallucinated categories during feature bank construction.

\subsection{Ablation Studies}
\label{sec:ablation}

\textbf{Effectiveness of Key Components.} All ablations use validation sets unless otherwise specified. \cref{tab:ablation_components} presents the quantitative ablation results. The baseline with C-RADIOv4 and EntitySeg achieves 84.3\% mIoU. Incorporating Soft-Masked Feature Aggregation (SMFA) alleviates boundary ambiguity during feature extraction and improves performance to 84.6\%. Semantic Boundary Purification (SBP) yields a substantial gain of 0.9\% by eroding uncertain regions to suppress feature mixing. The combination of both modules produces a synergistic effect, reaching a peak of 86.3\% mIoU. This confirms that purified boundaries paired with soft aggregation are essential for robust prototype learning. A per-category SBP analysis is provided in the supplementary material.

\begin{table}[!t]
    \centering
    \caption{\textbf{Ablation study of key components on VOC val set.} SBP and SMFA demonstrate synergistic effects for robust prototype learning.}
    \label{tab:ablation_components}
    \small
    \setlength{\tabcolsep}{12pt}
    \begin{tabular}{l|cc|c}
    \toprule
    Method & SBP & SMFA & mIoU \\
    \midrule
    Baseline & & & 84.3 \\
    w/ SMFA & & \checkmark & 84.6 \\
    w/ SBP & \checkmark & & 85.2 \\
    \textbf{ModuSeg} & \checkmark & \checkmark & \textbf{86.3} \\
    \bottomrule
    \end{tabular}
\end{table}

\textbf{Impact of Feature Bank Construction Strategies and Backbone Scalability.} 
\cref{tab:bank_construction} (Left) indicates that filtering is crucial for maintaining prototype integrity. Without it, background noise degrades mask quality for CorrCLIP and Mask Adapter. Our logits-based filtering removes this noise and improves mask mIoU. \cref{tab:bank_construction} (Right) validates the scalability of ModuSeg. Performance scales consistently with backbone strength from DINOv2 to C-RADIOv4, indicating that powerful pre-trained foundation representations are an important source of the gains and that our decoupled design leverages stronger backbones.

\begin{table}[!t]
    \centering
    \caption{\textbf{Impact of feature bank construction strategies and backbone scalability.} \textit{Left}: Effect of logits-based filtering on pseudo-mask mIoU over the training sets. \textit{Right}: Performance scaling with backbone strength.}
    \label{tab:bank_construction}
    \small
    \begin{minipage}{0.48\linewidth}
        \centering
        \setlength{\tabcolsep}{3pt}
        \begin{tabular}{l|l|cc} 
        \toprule
        Model & Filtering & \multicolumn{2}{c}{mIoU (\%)} \\
         & & VOC & COCO \\
        \midrule
        CorrCLIP & w/o Filter & 68.7 & 42.5 \\
         & \textbf{w/ Filter} & \textbf{78.8} & \textbf{54.6} \\
        \midrule
        Mask Adapter & w/o Filter & 76.6 & 55.1 \\
         & \textbf{w/ Filter} & \textbf{82.4} & \textbf{60.2} \\
        \bottomrule
        \end{tabular}
    \end{minipage}
    \hfill
    \begin{minipage}{0.48\linewidth}
        \centering
        \setlength{\tabcolsep}{4pt}
        \begin{tabular}{l|l|cc}
        \toprule
        Backbone & Size & \multicolumn{2}{c}{mIoU (\%)} \\
         & & VOC & COCO \\
        \midrule
        DINOv2 & ViT-B & 82.9 & 51.0 \\
        DINOv3 & ViT-B & 84.7 & 53.5 \\
        DINOv3 & ViT-L & 85.3 & 55.8 \\
        C-RADIOv4 & SO400M & \textbf{86.3} & \textbf{56.7} \\
        \bottomrule
        \end{tabular}
    \end{minipage}
\end{table}

\textbf{Scalability and Generalizability.} 
\cref{tab:scalability} demonstrates that ModuSeg's decoupled architecture can derive downstream segmentation gains from suitable pre-trained components while maintaining strong generalizability across different base models.
For mask proposals, \cref{tab:scalability} (Left) compares class-agnostic SAM 2 against EntitySeg. While SAM 2 excels at geometric delineation, EntitySeg demonstrates superior instance-level consistency, serving as a more suitable pre-trained geometric prior for our retrieval framework. 
\cref{tab:scalability} (Right) highlights ModuSeg's robustness across base models and datasets. The consistent gains on ADE20K and Cityscapes further support its generalizability across diverse datasets. ModuSeg also improves both correlation-based CorrCLIP and supervised Mask Adapter. Although Mask Adapter often yields stronger performance, our default configuration uses CorrCLIP to maintain a fully training-free pipeline with competitive results.

\begin{table}[!t]
    \centering
    \caption{\textbf{Scalability and generalizability of ModuSeg.} \textit{Left}: Comparison of mask proposal sources. \textit{Right}: Consistent plug-and-play gains across different models and datasets. ``Direct'' denotes zero-shot inference using the base model, while ``+ Ours'' indicates using the base model for pseudo-mask generation within our framework. ADE and City. denote ADE20K and Cityscapes, respectively.}
    \label{tab:scalability}
    \small
    \begin{minipage}{0.32\linewidth}
        \centering
        \setlength{\tabcolsep}{4pt}
        \begin{tabular}{l|cc}
        \toprule
        Source & \multicolumn{2}{c}{mIoU} \\
         & VOC & COCO \\
        \midrule
        SAM 2 & 82.6 & 54.8 \\
        \textbf{EntitySeg} & \textbf{86.3} & \textbf{56.7} \\
        \bottomrule
        \end{tabular}
    \end{minipage}
    \hfill
    \begin{minipage}{0.65\linewidth}
        \centering
        \setlength{\tabcolsep}{3pt}
        \newcommand{\ds}[1]{\makebox[2.5em][c]{#1}}
        \begin{tabular}{l|l|cccc}
        \toprule
        Model & Method & \ds{VOC} & \ds{COCO} & \ds{ADE} & \ds{City.} \\
        \midrule
        CorrCLIP & Direct & \ds{74.8} & \ds{45.0} & \ds{28.6} & \ds{51.6} \\
        & \textbf{+ Ours} & \ds{\textbf{86.3}} & \ds{\textbf{56.7}} & \ds{\textbf{44.7}} & \ds{\textbf{66.0}} \\
        \midrule
        Mask Adapter & Direct & \ds{81.2} & \ds{55.4} & \ds{38.2} & \ds{55.0} \\
        & \textbf{+ Ours} & \ds{\textbf{86.6}} & \ds{\textbf{59.3}} & \ds{\textbf{45.6}} & \ds{\textbf{64.2}} \\
        \bottomrule
        \end{tabular}
    \end{minipage}
\end{table}


\subsection{Deep Analysis}
\label{sec:deep_analysis}

\textbf{Hyper-parameter Analysis.} 
We provide comprehensive ablation studies of key hyper-parameters—including the Top-K retrieved neighbors, feature outlier removal ratio, and mask erosion iterations—in the supplementary material.

\begin{table}[!t]
    \centering
    \caption{\textbf{Internal property analysis.} \textit{Left}: The oracle analysis of bank quality and inference segmentation. \textit{Right}: Data efficiency with samples per category ($N$).}
    \label{tab:analysis_internal}
    \small
    \begin{minipage}{0.48\linewidth}
        \centering
        \setlength{\tabcolsep}{2pt}
        \begin{tabular}{l|cc|cc}
        \toprule
        Setting & Bank & Infer & \multicolumn{2}{c}{mIoU (\%)} \\
         & Source & Source & VOC & COCO \\
        \midrule
        Ours & Pseudo & Entity & \textbf{86.3} & \textbf{56.7} \\
        Oracle Bank & GT & Entity & 86.7 & 62.5 \\
        Oracle Seg & Pseudo & GT & 95.7 & 82.1 \\
        Oracle Sys & GT & GT & 95.7 & 84.9 \\
        \bottomrule
        \end{tabular}
    \end{minipage}
    \hfill
    \begin{minipage}{0.45\linewidth}
        \centering
        \setlength{\tabcolsep}{5pt}
        \begin{tabular}{l|cc}
        \toprule
        $N$-Shot & \multicolumn{2}{c}{mIoU (\%)} \\
        (Img/Cls) & VOC & COCO \\
        \midrule
        5 & 10.7 & 14.9 \\
        10 & 46.2 & 29.3 \\
        20 & 69.8 & 41.4 \\
        50 & 80.2 & 52.3 \\
        100 & 85.7 & 54.7 \\
        Full & \textbf{86.3} & \textbf{56.7} \\
        \bottomrule
        \end{tabular}
    \end{minipage}
\end{table}

\textbf{Performance Bounds.} 
\cref{tab:analysis_internal} (Left) reveals that substituting the pseudo-mask bank with ground truth annotations yields only marginal gains on VOC. This suggests that the quality of our retrieved features is comparable to that of supervised counterparts, implying that the feature bank is not the primary bottleneck. Conversely, employing ground truth during the inference segmentation stage substantially boosts performance to 95.7\%. This significant gap indicates that the limitation lies primarily in the class-agnostic segmenter, which lacks specific semantic guidance. Specifically, the segmenter often over-partitions complex backgrounds into discrete entities based on low-level geometric and textural cues. This behavior creates a substantial “semantic gap” compared to human-annotated ground truth, thereby hindering perfect alignment. Fully supervised counterparts are discussed in the supplementary material.

\textbf{Data Efficiency.} 
We further evaluate the data requirements in \cref{tab:analysis_internal} (Right). As shown, the model performance improves significantly as the number of samples ($N$) increases. Notably, the setting with only 50 shots per category retains approximately 93\% of the full-data performance on VOC. This demonstrates that our framework is highly data-efficient, effectively capturing the semantic distribution of categories even with limited support examples.

\textbf{Training Efficiency Comparison.}
\cref{tab:efficiency} compares the computational costs for model training. Unlike optimization-based methods like CLIMS and MCTformer+ that necessitate extensive GPU training, ModuSeg operates in a training-free manner without backpropagation. Consequently, ModuSeg achieves the lowest measured training time and the highest validation mIoU among the compared methods, highlighting the superior efficiency-performance trade-off of our retrieval-based paradigm. Online inference latency is further reported in the supplementary material.

\begin{table}[!t]
    \centering
    \caption{\textbf{Training efficiency comparison on VOC train set.} Type $\mathcal{M}$ denotes multi-stage methods and $\mathcal{S}$ denotes single-stage methods. All models are evaluated on RTX 3090.}
    \label{tab:efficiency}
    \small
    \setlength{\tabcolsep}{7pt}
    \begin{tabular}{l|c|ccc}
    \toprule
    Method & Type & Time (min) & GPU (G) & Val (mIoU) \\
    \midrule
    CLIMS {\scriptsize (CVPR'22)} \cite{xie2022clims}           & $\mathcal{M}$ & 1068 & 18.0 & 70.4 \\
    CLIP-ES {\scriptsize (CVPR'23)} \cite{lin2023clip}          & $\mathcal{M}$ & 420 & 12.0 & 72.2 \\
    MCTformer+ {\scriptsize (TPAMI'24)} \cite{xu2024mctformer+} & $\mathcal{M}$ & 1496 & 18.0 & 74.0 \\
    SeCo {\scriptsize (CVPR'24)} \cite{yang2024separate}        & $\mathcal{S}$ & 407 & 17.6 & 74.0 \\
    WeCLIP {\scriptsize (CVPR'24)} \cite{zhang2024frozen}       & $\mathcal{S}$ & 270 & 6.2 & 76.4 \\
    POT {\scriptsize (CVPR'25)}  \cite{wang2025pot}             & $\mathcal{M}$ & 678 & - & 76.1 \\
    \midrule
    \textbf{ModuSeg} & $\mathcal{M}$ & \textbf{84} & \textbf{5.3} & \textbf{86.3} \\
    \bottomrule
    \end{tabular}
\end{table}


\section{Conclusion}
\label{sec:conclusion}

In this paper, we propose ModuSeg, a training-free framework that explicitly decouples object discovery from semantic retrieval for weakly supervised semantic segmentation. By delegating localization to general mask proposals and constructing a robust offline feature bank, our approach circumvents the complexity of coupled optimization paradigms. Furthermore, our semantic boundary purification and soft-masked feature aggregation strategies effectively mitigate structural noise and quantization errors inherent in pseudo-labels. As demonstrated, ModuSeg achieves competitive performance and superior boundary adherence without parameter fine-tuning, offering a valuable perspective for leveraging foundation models in WSSS.

\section*{Acknowledgements}

This work was partially supported by the Guangdong Major Project of Basic and Applied Basic Research (2019B030302002), the National Natural Science Foundation of China (U24B20151), the Science and Technology Project of Guangdong Province (2021B1111600001), and the Major Key Project of PCL under Grants PCL2025A13 and PCL2025A08.


%
\bibliographystyle{splncs04}
\bibliography{main}
\end{document}